\documentclass[10pt,twocolumn,letterpaper]{article}

\usepackage{iccv}
\usepackage{times}
\usepackage{epsfig}
\usepackage{graphicx}
\usepackage{amsmath}
\usepackage{amssymb}
\usepackage{rotating}
\usepackage[table,xcdraw]{xcolor}
\usepackage{hhline}
\usepackage{multirow}
\usepackage{subcaption}
\usepackage[utf8x]{inputenc}
\usepackage{boldline}
\usepackage{graphicx}

\DeclareRobustCommand*{\IEEEauthorrefmark}[1]{%
  \raisebox{0pt}[0pt][0pt]{\textsuperscript{\footnotesize #1}}%
}

\usepackage[breaklinks=true,bookmarks=false]{hyperref}

\iccvfinalcopy % *** Uncomment this line for the final submission

\def\iccvPaperID{****} % *** Enter the ICCV Paper ID here

\ificcvfinal\fi

\begin{document}

%%%%%%%%% TITLE
\title{HarDNet: A Low Memory Traffic Network}
\author{
{Ping Chao\IEEEauthorrefmark{1,2} 
\quad Chao-Yang Kao\IEEEauthorrefmark{1} 
\quad Yu-Shan Ruan\IEEEauthorrefmark{1} 
\quad Chien-Hsiang Huang\IEEEauthorrefmark{1}
\quad Youn-Long Lin\IEEEauthorrefmark{1}}\\
\IEEEauthorrefmark{1}National Tsing Hua University \quad \IEEEauthorrefmark{2}University of Michigan\\
{\tt\small pingchao@umich.edu 
\quad \{chaoyangkao923, esunxd, james128333\}gmail.com 
\quad ylin@cs.nthu.edu.tw}
}

\maketitle 

\ificcvfinal\thispagestyle{empty}\fi

\begin{abstract}
   State-of-the-art neural network architectures such as ResNet, MobileNet, and DenseNet have achieved outstanding accuracy over low MACs and small model size counterparts. However, these metrics might not be accurate for predicting the inference time. We suggest that memory traffic for accessing intermediate feature maps can be a factor dominating the inference latency, especially in such tasks as real-time object detection and semantic segmentation of high-resolution video. We propose a Harmonic Densely Connected Network to achieve high efficiency in terms of both low MACs and memory traffic. The new network achieves 35\%, 36\%, 30\%, 32\%, and 45\% inference time reduction compared with FC-DenseNet-103, DenseNet-264, ResNet-50, ResNet-152, and SSD-VGG, respectively. We use tools including Nvidia profiler and ARM Scale-Sim to measure the memory traffic and verify that the inference latency is indeed proportional to the memory traffic consumption and the proposed network consumes low memory traffic. We conclude that one should take memory traffic into consideration when designing neural network architectures for high-resolution applications at the edge.
\end{abstract}

%%%%%%%%% BODY TEXT
\section{Introduction}
Convolutional Neural Networks (CNN) have been popular for computer vision tasks, ever since the explosive growth of computing power has made possible training complex networks like AlexNet\cite{A.Krizhevsky, A.Krizhevsky2}, VGG-net\cite{K.Simonyan}, and Inception\cite{C.Szegedy} in a reasonable amount of time. 
To bring these fascinating research results into mass use, performing a neural network inference on edge devices is inevitable.
However, edge computing relies on limited computation power and battery capacity. How to increase computation efficiency and reduce the power consumption for neural network inference at the edge has therefore become a critical issue.

Reducing model sizes (the number of parameters or weights of a model) is a hot research topic in improving both computation and energy efficiency, since a reduced model size usually implies fewer MACs (number of multiply-accumulate operations or floating point operations) and less dynamic random-access memory (DRAM) traffic for read and write of model parameters and feature maps. 
Several researches have steered toward maximizing the accuracy–parameters ratio. 
State-of-the-art networks such as Residual Networks (ResNets)\cite{K.He}, SqueezeNets\cite{F.N.Iandola}, and Densely Connected Networks (DenseNets)\cite{G.Huang} have achieved high parameter efficiency that have dramatically reduced the model size while maintaining a high accuracy. The model size can be reduced further through compression. 
Han et al.\cite{S.Han} showed that the large amount of floating-point weights loaded from DRAM may consume more power than arithmetic operations do. 
Their Deep Compression algorithm employs weight pruning and quantization to reduce the model size and power consumption significantly.\\

In addition to the power consumption, DRAM accesses can also dominate system performance in terms of inference time due to the limited DRAM bandwidth. 
Since we have observed that the size summation of all the intermediate feature maps in a CNN can be ten to hundred times larger than its model size, especially for high resolution tasks such as semantic segmentation using fully convolutional networks\cite{J.Long}, we suggest that reducing DRAM accesses to feature maps may lead to a speedup in some cases.\\

Shrinking the size of feature maps is a straightforward approach to reduce the traffic. While there are only a few papers addressing lossless compression of feature maps, lossy compression of feature maps has been intensively studied in research of model precision manipulation and approximation\cite{A.Aimar, M.Courbariaux, P.Gysel, D.Miyashita, M.Rastegari}. 
The quantization used in these works for model compression can usually reduce the feature map size automatically. However, like other lossy compression methods such as subsampling, they usually penalize accuracy. In this paper, we explore how to reduce the DRAM traffic for feature maps without penalizing accuracy simply by designing the architecture of a CNN carefully.\\

To design such a low DRAM traffic CNN architecture,
it is necessary to measure the actual traffic. 
For a general-purpose Graphics Processing Unit (GPU),
we use Nvidia profiler to measure
the number of DRAM read/write bytes. 
For mobile devices, we use ARM Scale Sim \cite{samajdar2018scale} to get traffic data and inference cycle counts for each CNN architecture. 
We also propose a metric called Convolutional Input/Output (CIO), which is simply a summation of the input tensor size and output tensor size of every convolution layer as equation (1), where \(c\) is the number of channels and \(w\) and \(h\) are the width and height of the feature maps for a convolution layer \textit{l}.

\begin{equation}
CIO = \sum_{l}(c_{in}^{(l)}\times w_{in}^{(l)}\times h_{in}^{(l)} + c_{out}^{(l)}\times w_{out}^{(l)}\times h_{out}^{(l)})
\end{equation}

CIO is an approximation of DRAM traffic proportional to the real DRAM traffic measurement. Please note that the input tensor can be a concatenation, and a reused tensor can therefore be counted multiple times. Using a lot of large convolutional kernels may easily achieve a minimized CIO. However, it also damages the computational efficiency and eventually leads to a significant latency overhead outweighing the gain. Therefore, we argue that maintaining a high computational efficiency is still imperative, and CIO dominates the inference time only when the computational density, which is, the MACs over CIO (MoC) of a layer, is below a certain ratio that depends on platforms. \\

For example, under a fixed CIO, changing the channel ratio between the input and output of a convolutional layer step by step from 1:1 to 1:100 leads to reductions of both MACs and latency. For the latency, it declines more slowly than the reduction of MACs, since the memory traffic remains the same. A certain value of MoC may show that, below this ratio, the latency for a layer is always bounded to a fixed time. However, this value is platform-dependent and obscure empirically.\\

\begin{figure}[t]
\begin{center}
\includegraphics [width=0.9\linewidth]{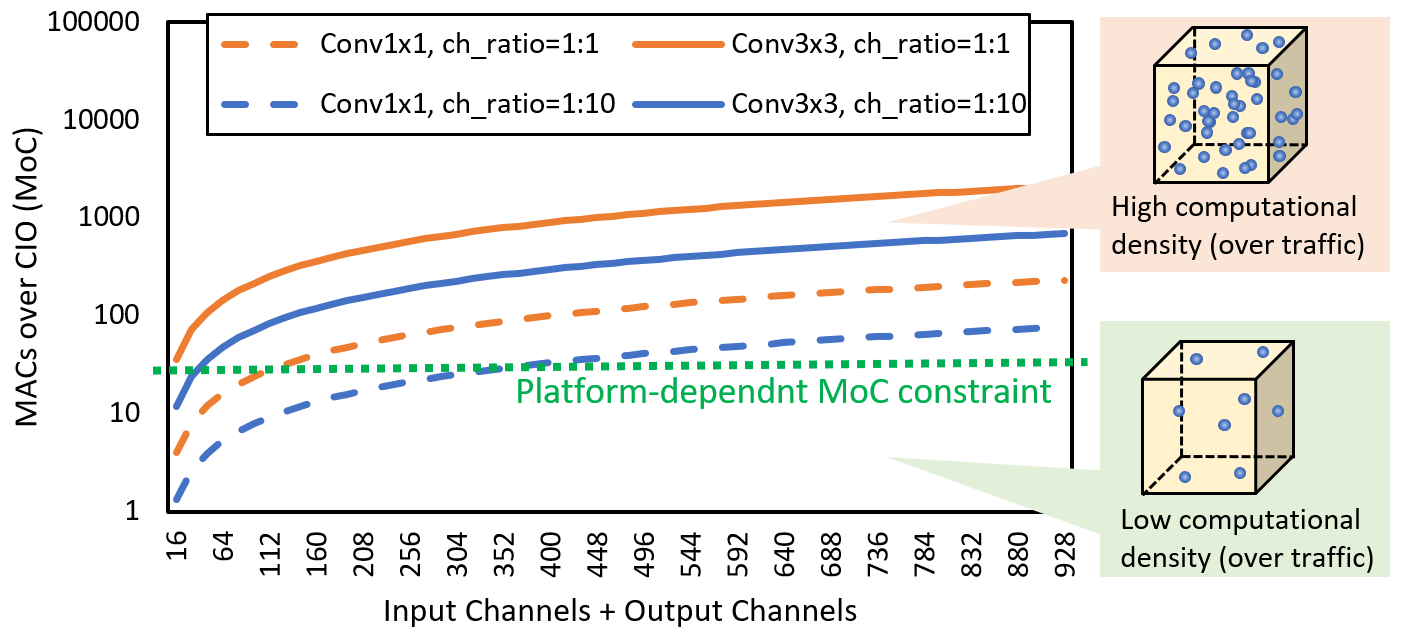}

\end{center}
\caption {Concept of MoC constraint. A Conv layer with MoC below the constraint is avoided. }
\label{figure:fig0}
\end{figure}

In this paper, we apply a soft constraint on the MoC of each layer to design a low CIO network model with a reasonable increase of MACs. As shown in Fig. \ref{figure:fig0}, we avoid to employ a layer with a very low MoC such as a Conv1x1 layer that has a very large input/output channel ratio. Inspired by the Densely Connected Networks \cite{G.Huang} we propose a Harmonic Densely Connected Network (HarDNet) by applying the strategy. We first reduce most of the layer connections from DenseNet to reduce concatenation cost. Then, we balance the input/output channel ratio by increasing the channel width of a layer according to its connections.\\

The contribution of this paper is that we introduce DRAM traffic for feature map access and its platform-independent approximation, CIO, as a new metric for evaluating a CNN architecture and show that the inference latency is highly correlated with the DRAM traffic. By constraining the MoC of each layer, we propose HarDNets that reduces DRAM traffic by 40\% compared with DenseNets. 
We evaluate the proposed HarDNet on the CamVid\cite{Brostow}, ImageNet (ILSVRC)\cite{J.Deng}, PASCAL VOC\cite{pascal}, and MS COCO \cite{coco} datasets. Compared to DenseNet and ResNet, HarDNet achieves the same accuracy with 30\%\(\sim\)50\% less CIO, and accordingly, 30\%\(\sim\)40\% less inference time.

%-------------------------------------------------------------------------
\section{Related works}
A significant trend in neural network research is exploiting shortcuts. To cope with the degradation problem, Highway Networks\cite{R.K.Srivastava} and Residual Networks\cite{K.He} add shortcuts to sum up a layer with multiple preceeding layers. The stochastic depth regularization\cite{G.Huang2} is essentially another form of shortcuts for crossing layers that are randomly dropped. Shortcuts enable implicit supervision to make networks continually deeper without degradation. DenseNets\cite{G.Huang} concatenates all preceeding layers as a shortcut achieving more efficient deep supervision. Shortcuts have also been shown to be very useful in segmentation tasks\cite{M.Drozdzal}. Jégou et al.\cite{FCD} showed that without any pre-training, DenseNet performs semantic segmentation very well. However, shortcuts lead to both large memory usage and heavy DRAM traffic. Using shortcuts elongates the lifetime of a tensor, which may result in frequent data exchanges between DRAM and cache.\\

Some sparsified versions of DenseNet have been proposed. LogDenseNet\cite{H.Z.Hu} and SparseNet\cite{L.G.Zhu} adopt a strategy of sparsely connecting each layer \(k\) with layer \(k – 2^n\) for all integers \(n ≥  0\) and \(k – 2^n ≥ 0\) such that the input channel numbers decrease from \(O(L^2)\) to \(O(L \log L)\). The difference between them is that LogDenseNet applies this strategy globally, where layer connections crossing blocks with different resolutions still follow the log connection rule, while SparseNet has a fixed block output that regards the output as layer \(L+1\) for a block with \(L\) layers. However, both network architectures need to significantly increase the growth rate (output channel width) to recover the accuracy dropping from the connection pruning, and the increase of growth rate can compromise the CIO reduction. Nevertheless, these studies did point out a promising direction to sparsify the DenseNet.\\

The performance of a classic microcomputer architecture is dominated by its limited computing power and memory bandwidth\cite{mem}. Researchers focused more on enhancing the computation power and efficiency. Some researchers pointed out that limited memory bandwidth can dominate the inference latency and power consumption in GPU-based systems \cite{C.Li, J.Long}, FPGA-based systems \cite{S.Chakradhar, C.Farabet}, or custom accelerators\cite{A.Aimar, T.Chen, Y.Chen}. However, there is no systematic way to correlate DRAM traffic and the latency. Therefore, we propose CIO and MoC and present a conceptual methodology for enhancing efficiency.

%-------------------------------------------------------------------------
\section{Proposed Harmonic DenseNet}
\subsection{Sparsification and weighting}
\begin{figure}[t]
\begin{center}
\includegraphics [width=1.0\linewidth]{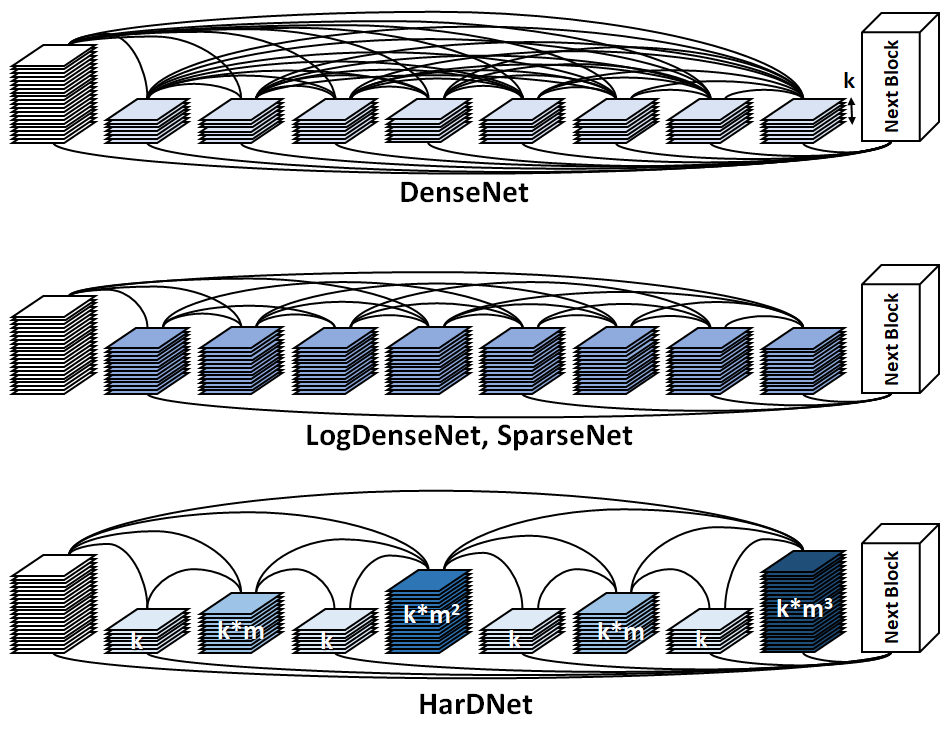}
\end{center}
\caption {Illustrations for DenseNet, LogDenseNet, SparseNet, and the proposed Harmonic DenseNet (HarDNet), in which each of the layers is a 3x3 convolution. } 
\label{figure:fig1}
\end{figure}

We propose a new network architecture based on the Densely Connected Network. Unlike the sparsification proposed in LogDenseNet, we let layer \(k\) connect to layer \(k – 2^n\) if \(2^n\) divides \(k\), where \(n\) is a non-negative integer and \(k – 2^n ≥ 0\); specifically, layer 0 is the input layer. Under this connection scheme, once layer \(2^n\) is processed, layer \(1\) through \(2^n – 1\) can be flushed from the memory. The connections make the network appear as an overlapping of power-of-two-th harmonic waves, as illustrated in Fig. \ref{figure:fig1}, hence we name it the Harmonic Densely Connected Network (HarDNet). The proposed sparsification scheme reduces the concatenation cost significantly better than the LogDenseNet does. This connection pattern also looks like a FractalNet \cite{G.Larsson}, except the latter uses averaging shortcuts instead of concatenations.\\

In the proposed network, layers with an index divided by a larger power of two are more influential than those that divided by a smaller power of two. We amplify these key layers by increasing their channels, which can balance the channel ratio between the input and output of a layer to avoid a low MoC. A layer \(l\) has an initial growth rate \(k\), and we let its channel number to be \(k \times m^n\), where \(n\) is the maximum number satisfying that \(l\) is divided by \(2^n\). The multiplier m serves as a low-dimensional compression factor. If the input layer 0 has \(k\) channels and \(m=2\), we get a channel ratio 1:1 for every layer. Setting \(m\) smaller than two is tantamount to compress the input channels into fewer output channels. Empirically, setting \(m\) between 1.6 and 1.9 achieves a good accuracy and parameter efficiency.

\subsection{Transition and Bottleneck Layers}
The proposed connection pattern forms a group of layers called a Harmonic Dense Block (HDB), which is followed by a Conv1x1 layer as a transition. We let the depth of each HDB to be a power of two such that the last layer of an HDB has the largest number of channels. In DenseNet, a densely connected output of a block directly passes the gradient from output to all preceding layers to achieve deep supervision. In our HDB with depth \(L\), the gradient will pass through at most \(\log L\) layers. To alleviate the degradation, we made the output of a depth-\(L\) HDB to be the concatenation of layer \(L\) and all its preceeding odd numbered layers, which are the least significant layers with \(k\) output channels. The output of all even layers from 2 to \(L–2\) can be discarded once the HDB is finished. Their total memory occupation is roughly two to three times as large as all the odd layers combined when \(m\) is between 1.6 to 1.9.\\

%-------------------------------------------------------------------------

DenseNet employees a bottleneck layer before every Conv3x3 layer to enhance the parameter efficiency. Since we have balanced the channel ratio between the input and output for every layer, the effect of such bottleneck layers became insignificant. Inserting a bottleneck layer for every four Conv3x3 layer is still helpful for reducing the model size. We let the output channels of a bottleneck layer to be \(\sqrt{c_{in} / c_{out}} \times c_{out}\), where \(c_{in}\) is the concatenated input channels and \(c_{out}\) is the output channels of the following Conv3x3 layer. To further improve the inference time, these Conv1x1 bottleneck layers can be discarded to meet our MoC constraint.\\

\begin{figure}	
	\centering
	\begin{subfigure}[h]{1.2in}
		\centering
		\includegraphics[width=1.1in]{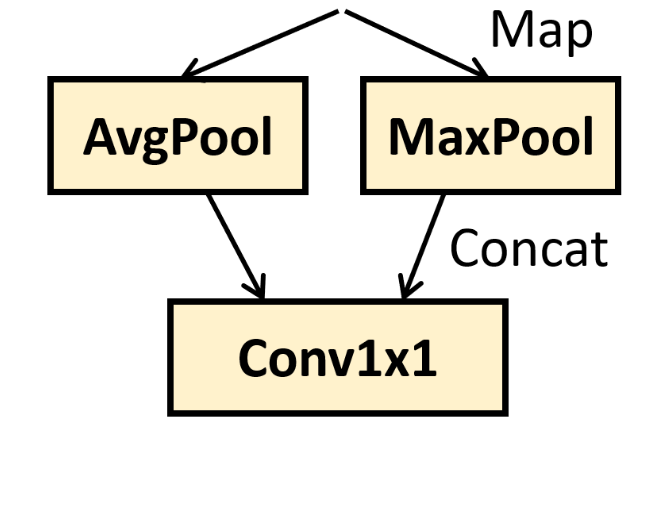}
		\caption{}		
		\label{fig:fig2-2a}
	\end{subfigure}
	\quad
	\begin{subfigure}[h]{1.2in}
		\centering
		\includegraphics[width=1.1in]{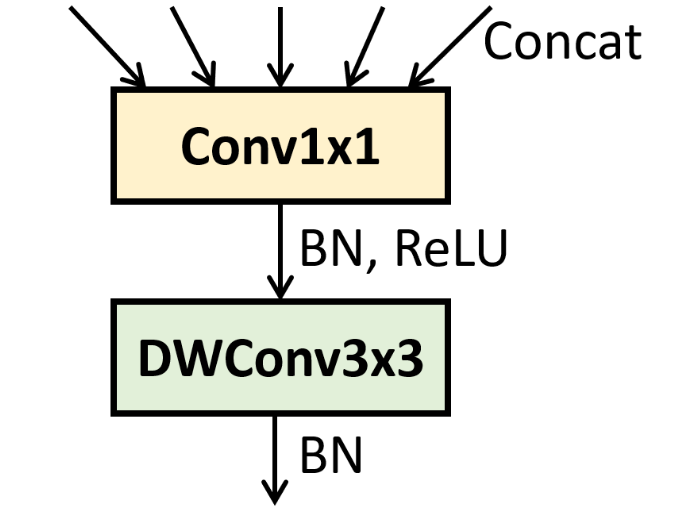}
		\caption{}
		\label{fig:fig2-2b}
	\end{subfigure}
	\caption{ (a) Inverted transition down module, (b)Depthwise-separable convolution for HarDNet}\label{figure:fig2-2}
\end{figure}

The transition layer proposed by DenseNet is a Conv1x1 layer followed by a 2x2 average pooling. As shown in Fig. \ref{fig:fig2-2a}, we propose an inverted transition module, which maps input tensor to an additional max pooling function along with the original average pooling, followed by concatenation and Conv1x1. This module reduces 50\% of CIO for the Conv1x1 while achieving roughly the same accuracy at the expense of model size increase.

\subsection{Detailed Design}

To compare with DenseNet, we follow its global dense connection strategy that bypasses all the input of an HDB as a part of its output and propose six models of HarDNet. The detailed parameters are shown in Table 1. We use a 0.85 reduction rate for the transition layers instead of the 0.5 reduction rate used in the DenseNet, since a low-dimensional compression has been applied to the growth rate multiplier as we mentioned before. To achieve a flexible depth, we partition a block into multiple blocks with 16 layers (20 when bottleneck layers are counted).\\

\newcommand{\bigcell}[2]{\begin{tabular}{@{\hspace{0cm}}#1@{}}#2\end{tabular}}

\renewcommand{\arraystretch}{1.1}
\begin{table}[t]
\scalebox{0.83}{

\begin{tabular}{ c | c | c | c | c | c }
 \hlineB{4}
 & 96s/L & 117s/L & 138s/L & 68 & 39DS\\[2ex]
 \hline
 k & 20/26 & 26/30 & 30/32 & \multicolumn{2}{c}{-}\\
 \hline
 m & \multicolumn{2}{c|}{1.6} & 1.6/1.65 & 1.7 & 1.6\\
 \hline
 red & \multicolumn{3}{c|}{0.85} & \multicolumn{2}{c}{-} \\
 \hline
 bottleneck & \multicolumn{3}{c|}{Y} & \multicolumn{2}{c}{N}\\
 \hlineB{4}
\multirow{2}{*}{Stride 2} & \multicolumn{3}{c|}{7x7, 64, stride=2} & \bigcell{c}{3x3, 32,\\ stride=2} & \bigcell{c}{3x3, 24,\\ stride=2}\\
 \cline{5-6}
 & \multicolumn{3}{c|}{} & 3x3, 64 & 1x1, 48\\
 \hline
 Stride 4 & \bigcell{c}{8 (HDB\\ depth)} & 8 & 8 & \bigcell{c}{8, k=14\\ t=128} & \bigcell{c}{4, k=16\\ t=96} \\
 \hline
 \multirow{2}{*}{Stride 8} &  \multirow{2}{*}{16} & \multirow{2}{*}{16} & \multirow{2}{*}{16} &\bigcell{c}{ 16, k=16\\ t=256} & \bigcell{c}{16, k=20\\ t=320} \\
 \cline{5-6}
 & & & & \bigcell{c}{16, k=20\\ t=320} & - \\
 \hline
 Stride 16 & 16\(\times\)2 & 16\(\times\)3 & 16\(\times\)3 & \bigcell{c}{16, k=40\\ t=640} & \bigcell{c}{8, k=64\\ t=640}\\
 \hline
 Stride 32 & 16 & 16 & 16\(\times\)2 & \bigcell{c}{4, k=160\\ t=1024} & \bigcell{c}{4, k=160\\ t=1024}\\
 \hlineB{4}
\end{tabular}

}
\caption{Detailed implementation parameters. A “3x3, 64” stands for a Conv3x3 layer with 64 output channels, and the leading numbers below Stride 2 stand for an HDB with how many layers, followed by its growth rate \(k\) and a transitional Conv1x1 with \(t\) output channels.}
\label{table:table0}
\end{table}

We further propose a HarDNet-68, in which we remove the global dense connections and use MaxPool for down-sampling, and we change the BN-ReLU-Conv order proposed by DenseNet into the standard order of Conv-BN-ReLU to enable the folding of batch normalization. The dedicated growth rate k for each HDB in the HarDNet-68 enhances the CIO efficiency. Since a deep HDB has a larger number of input channels, a larger growth rate helps to balance the channel ratio between the input and output of a layer to meet our MoC constraint. For the layer distribution, instead of concentrating on stride-16 that is adopted by most of the CNN models, we let stride-8 to have the most layers in the HarDNet-68 that improves the local feature learning benefiting small-scale object detection. In contrast, classification tasks rely more on the global feature learning, so concentrating on the low resolution achieves a higher accuracy and a lower computational complexity.\\

The depth separable convolution that dramatically reduces model size and computational complexity is also adoptable on the HarDNet. We propose a HarDNet-39DS with pure depth-wise-separable (DS) convolutions except the first convolutional layer by decomposing a Conv3x3 layer into a point-wise convolution and a depth-wise convolution as shown in Fig. \ref{fig:fig2-2b}. The order matters in this case. Since every layer in an HDB has a wide input and a narrow output, inverting the order increases the CIO dramatically. Please note that CIO may not be a direct prediction of inference latency for the comparison between a model with standard Conv3x3 and a model with depth-wise separable convolutions, because there is a huge difference of MACs between them. Nevertheless, the prediction can still be achieved when there is a weighting applied on the CIO for the decomposed convolution.

\section{Experiments}

\begin{table*}[ht!]
\scalebox{0.79}{\begin{tabular}{|l|c|c|c|c||c|c|c|c|c|c|c|c|c|c|c||c|c|}
\hline
\multicolumn{1}{|c|}{\rotatebox{90}{Method}}  & \rotatebox{90}{\begin{tabular}[c]{@{}l@{}}GMACs,\\ @352x480\end{tabular}} & \rotatebox{90}{\# Params (M)} & \rotatebox{90}{CIO,(MB)} & \rotatebox{90}{GPU time (s)} & \rotatebox{90}{Building} & \rotatebox{90}{Tree} & \rotatebox{90}{Sky} & \rotatebox{90}{Car} & \rotatebox{90}{Sign} & \rotatebox{90}{Road} & \rotatebox{90}{Pedestrian} & \rotatebox{90}{Fence} & \rotatebox{90}{Pole} & \rotatebox{90}{Sidewalk} & \rotatebox{90}{Cyclist} &\rotatebox{90}{Mean IoU} & \rotatebox{90}{Global Acc.} \\\hhline{-----||-----------||--}
SegNet\cite{segnet} &224&29.5&702&3.7&68.7&52.0&87.0&58.5&13.4&86.2&25.3&17.9&16.0&60.5&24.8&46.4&62.5\\ 
\rowcolor[HTML]{E0FFFF} 
FCN8\cite{J.Long} &143&135&318&4.9&77.8&71.0&88.7&76.1&32.7&91.2&41.7&24.4&19.9&72.7&31.0&57.0&88.0\\
FC-DenseNet56\cite{FCD} &60&1.4&1351&6.1&77.6&72.0&92.4&732.&31.8&92.8&37.9&26.2&32.6&79.9&31.1&58.9&88.9\\
\rowcolor[HTML]{E0FFFF} 
FC-DenseNet67\cite{FCD} &140&3.5&2286&10.2&80.2&75.4&93.0&78.2&40.9&94.7&58.4&30.7&38.4&81.9&52.1&65.8&90.8\\
FC-DenseNet103\cite{FCD} &134&9.4&2150&11.4&83.0&77.3&93.0&77.3&43.9&94.5&59.6&37.1&37.8&82.2&50.5&66.9&91.5\\ 
\rowcolor[HTML]{E0FFFF} 
LogDenseNet-103\cite{H.Z.Hu} &137&4.7&2544&-&81.6&75.5&92.3&81.9&44.4&92.6&58.3&42.3&37.2&77.5&56.6&67.3&90.7\\ \hline\hline
FC-DenseNet-ref100&142&3.5&3337&15.2&81.1&77.1&92.9&77.7&40.8&94.3&58.1&35.2&37.0&81.5&48.9&65.8&90.9\\ 
\rowcolor[HTML]{E0FFFF} 
FC-SparseNet-ref100&223&3.2&2559&11.8&83.3&78.3&93.3&78.9&42.5&94.5&57.5&33.1&41.6&82.9&46.9&66.6&91.7\\
FC-HarDNet-ref100&151&3.6&2076&10&82.6&75.5&92.8&78.3&43.2&95.4&59.2&34.9&38.9&85.1&52.6&67.1&91.7\\ \hline\hline
\rowcolor[HTML]{E0FFFF} 
FC-HarDNet68&15&1.4&473&3.1&80.8&74.4&92.7&76.1&40.6&93.3&47.9&29.3&33.3&78.3&45.7&62.9&90.2\\
FC-HarDNet76&54&3.5&932&4.9&82.0&75.8&92.7&76.8&42.6&94.7&58.0&30.9&37.6&83.2&49.9&65.8&91.2\\ 
\rowcolor[HTML]{E0FFFF} 
FC-HarDNet84&100&8.4&1267&6.7&81.4&76.2&92.9&78.3&48.9&94.6&61.9&37.9&38.2&80.5&54.0&67.7&91.1\\ \hline
\end{tabular}
}
\caption{Results on CamVid dataset. The GPU inference time results are the accumulated measurements of CamVid test-set (233 pics) with a single-image batch size, running on pytorch-1.0.1 framework with a single NVIDIA TitanV GPU.}
\label{table:table1}
\end{table*}

\subsection{CamVid Dataset}

To study the performance of HDB, we replace all the blocks in a FC-DenseNet with HDBs. We follow the architecture of FC-DenseNet with an encoder-decoder structure and block level shortcuts to create models for semantic segmentation. For fair comparison, we made two reference architectures with exactly the same depth for each block and roughly the same model size and MACs, named FC-HarDNet-ref100 and FC-DenseNet-ref100, respectively. We trained and tested both networks on the CamVid dataset with 800 epochs and 0.998 learning rate decay on exactly the same environments, and followed the batch sizes of the two passes used in the original work\cite{FCD}. Table \ref{table:table1} shows the experiment results in mean IoU of both overall and per-classes. Comparing these two networks, FC-HarDNet-ref100 achieved a higher mean IoU and 38\% less CIO. When running inference testing on a single NVIDIA TitanV GPU, we observed 24\% and 36\% inference time savings using tensorflow and Pytorch frameworks, respectively. Since FC-HarDNet-ref100 consumes slightly more MACs than FC-DenseNet-ref100 does, the inference time saving should come from the memory traffic reduction.\\

\begin{table}[b]
\scalebox{0.78}{
\begin{tabular}{l|c|c|c|c}\hlineB{3.5}
&\begin{tabular}[c]{@{}l@{}}1st\\ Conv\end{tabular}& BLK depth  & Growth Rate & m\\ \hline
FC-D 103    & 48& 4, 5, 7, 10 , 12, 15 & 16 & -  \\ \hline
FC-D ref100 & 48& 8,  8, 8, 8, 8, 8 & 10 & -    \\ \hline
FC-S ref100 & 48& 8, 8, 8, 8, 8, 8& 26 & -    \\ \hline
FC-H ref100 & 48& 8, 8, 8, 8, 8, 8  & 10 & 1.54 \\ \hline
FC-H 68 & 8& 4, 4, 4, 4, 8, 8     & 4, 6, 8, 8, 10, 10 & 1.7  \\ \hline
FC-H 76& 24  & 4, 4, 4, 8, 8, 8     & 8,10,12,12,12,14   & 1.7 \\ \hline
FC-H 84& 32& 4, 4, 8, 8, 8, 8& 10,12,14,16,20,22  & 1.7 \\ \hlineB{3.5}
\end{tabular}
}
\caption{Parameters of FC-HarDNet and other reference networks, where FC-D, FC-S, and FC-H stand for FC-DenseNet, FC-SparseNet, and FC-HarDNet, respectively.}
\label{table:table2}
\end{table}

\begin{figure*}[ht]
    \centering
    \begin{subfigure}[b]{0.3\linewidth}
        \includegraphics[width=0.93\linewidth]{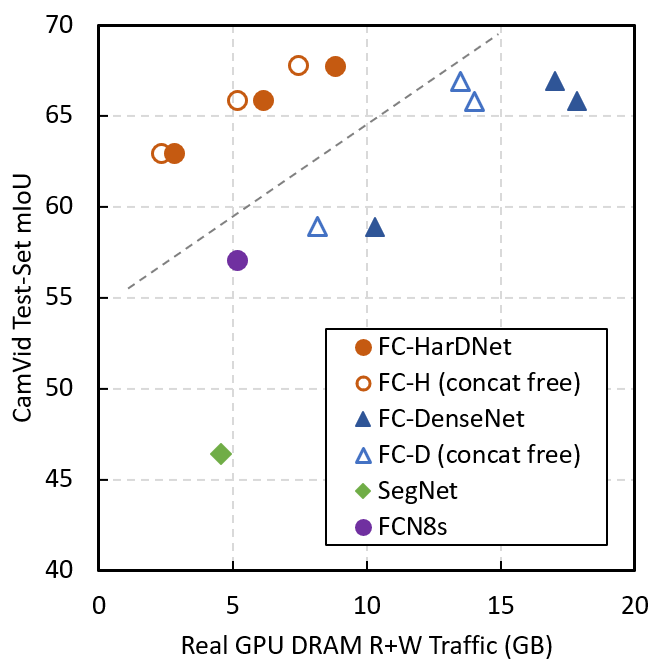}
        \caption{}
        \label{fig:fig2a}
    \end{subfigure}
    \begin{subfigure}[b]{0.36\linewidth}
        \includegraphics[width=0.99\linewidth]{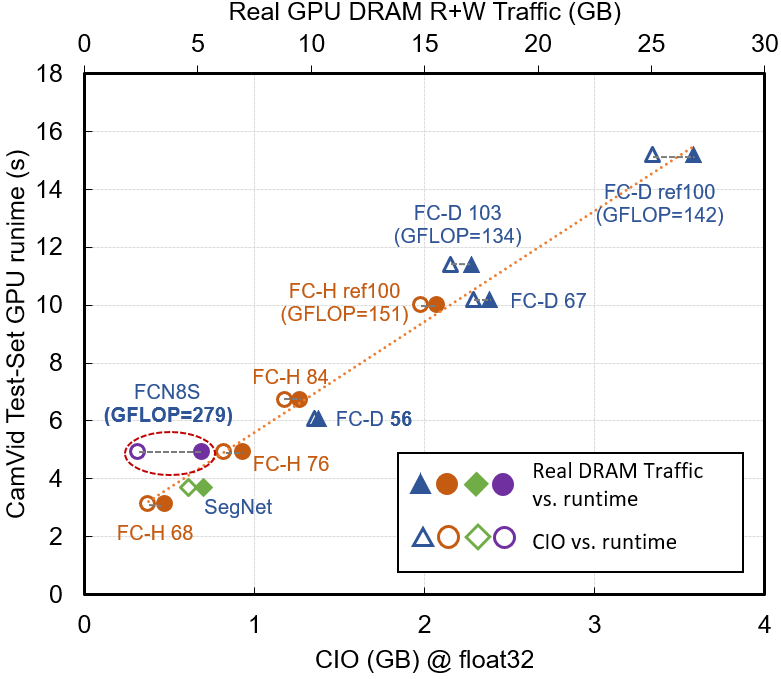}
        \caption{}
        \label{fig:fig2b}
    \end{subfigure}
    \begin{subfigure}[b]{0.31\linewidth}
        \includegraphics[width=1\linewidth]{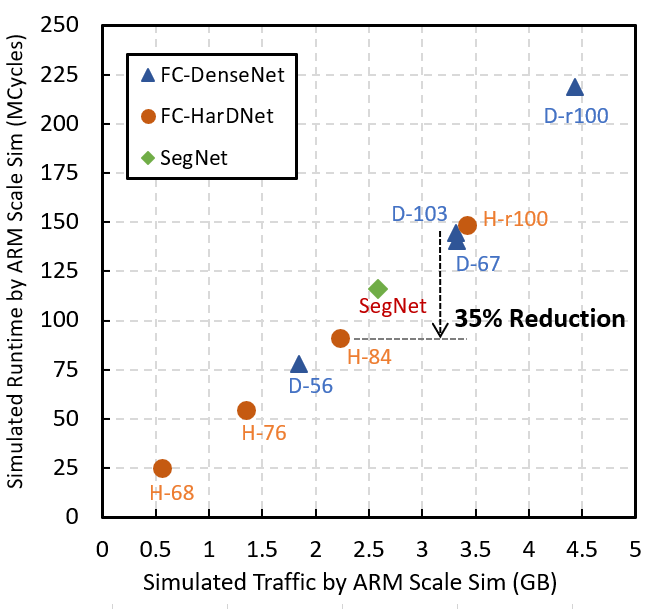}
        \caption{}
        \label{figure:fig3}
    \end{subfigure}
    
    \caption{ Correlation among accuracy, DRAM traffic, and GPU inference time for CamVid test set @ 360x480 running on a Nvidia Titan V with Cuda 9.0. (a) Mean IoU vs. DRAM traffic measured by Nvidia Profiler, where the concat-free sets stand for the case if the explicit memory copy for tensor concatenation can be completely removed. The two reference networks are not compared. (b) GPU inference time vs. DRAM traffic and CIO on Pytorch 1.0.1 framework. (c) Inference time vs. DRAM traffic measured by the simulation of Scale Sim.}
\label{fig:visual_smap}
\end{figure*}

Compared with other sparsified versions of DenseNet, Table \ref{table:table1} shows that FC-LogDenseNet103 gets a worse CIO number than the FC-DenseNet103 due to the long lifetime of the first half of layers caused by its global transition. On the other hand, SparseNets uses a localized transition layer such that it can reduce the tensor lifetime better than LogDenseNet. Therefore, we implemented a FC-SparseNet-ref100 for comparison and trained it in the same environment for five runs, and then we picked the best result. The result shows that FC-SparseNet can also reduce GPU inference time, but not as much as FC-HarDNet-ref100 does.\\

We propose FC-HarDNet84 as specified in Table \ref{table:table2} for comparing with FC-DenseNet103. The new network achieves CIO reduction by 41\% and GPU inference time reduction by 35\%. A smaller version, FC-HarDNet68, also outperforms FC-DenseNet56 by a 65\% less CIO and 52\% less GPU inference time. We investigated the correlations among accuracy, DRAM traffic, and GPU inference time. Fig. \ref{fig:fig2a} shows that HarDNet achieves the best accuracy-over-DRAM-traffic than other networks. Fig. \ref{fig:fig2b} shows that GPU inference time is indeed correlated with DRAM traffic much more than MACs. It also shows that CIO is a good approximation to the real DRAM traffic, except that FCN8s is an outlier due to its use of large convolutional kernels.\\

 To verify the correlation between inference time and memory traffic on hardware platforms differ from GPU, we employ ARM Scale Sim for the investigation. It is a cycle-accurate simulation tool for ARM’s systolic array or Eyeriss. Note that this tool does not support deconvolution and regards these deconv layers as ordinary convolutional layers. Fig. \ref{figure:fig3} shows that the correlation between DRAM traffic and inference time on the Scale Sim is still high, and FC-HarDNet-84 still reduces inference time by 35\% compared to FC-DenseNet-103. However, it also shows that the relative inference time of SegNet is much worse than on GPU. Thus, it confirmed that the relative DRAM traffic can be very different among platforms.\\

Pleiss et al. have mentioned that there is a concatenation overhead with the DenseNet implementation, which is caused by the explicit tensor copy from existing tensors to a new memory allocation. Therefore, it causes an additional DRAM traffic. To show that HarDNet still outperforms DenseNet when the overhead is discounted, we subtract the measured DRAM traffic volume by the traffic for tensor concatenation as the concat-free cases shown in Fig. \ref{fig:fig2a}, where the DRAM traffic of concatenation is measured by Nvidia Profiler and broken down to the CatArrayBatchedCopy function. Fig. \ref{fig:fig2a} shows that FC-DenseNet can reduce more DRAM traffic by discounting the concatenation than that for FC-HarDNet, but the latter still outperforms the former.

\begin{figure*}

    \begin{subfigure}[b]{0.5\textwidth}
        \begin{center}
        \includegraphics[width=0.8\textwidth]{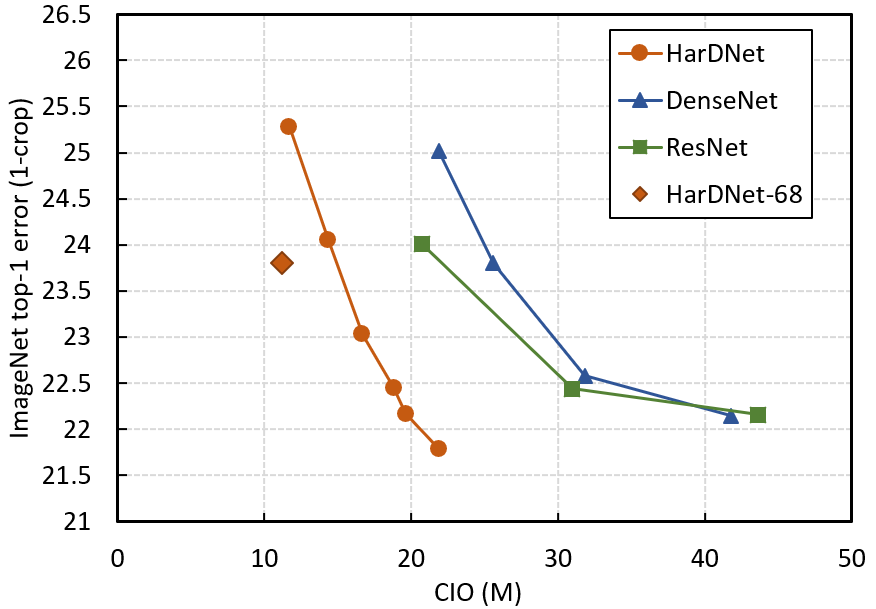}
        \caption{}
        \label{fig:fig5a}
        \end{center}
    \end{subfigure}
    \begin{subfigure}[b]{0.5\textwidth}
        \begin{center}
        \includegraphics[width=0.8\textwidth]{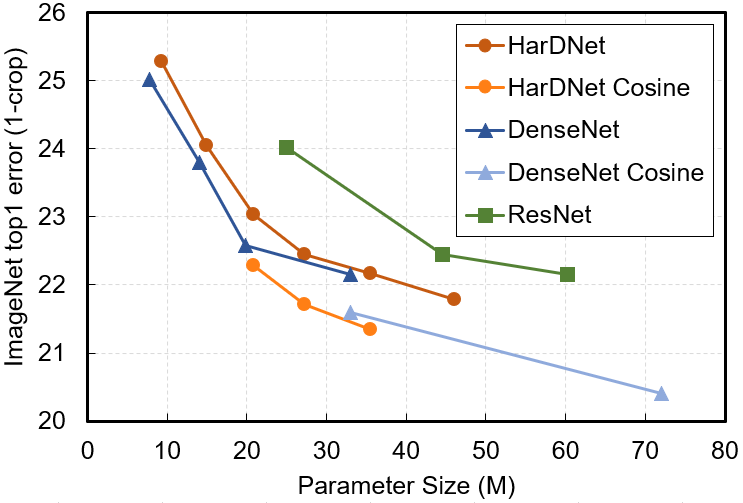}
        \caption{}
        \label{fig:fig5b}
        \end{center}
    \end{subfigure}
    \begin{subfigure}[b]{0.5\textwidth}
        \begin{center}
        \includegraphics[width=0.8\textwidth]{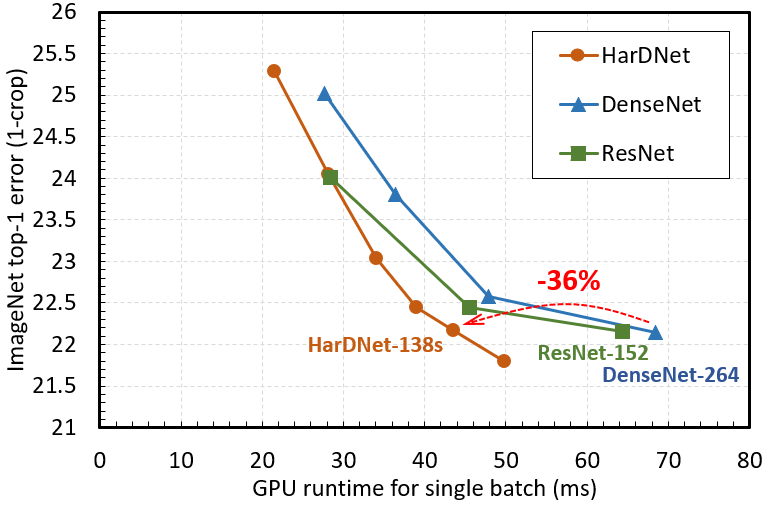}
        \caption{}
        \label{fig:fig5c}
        \end{center}
    \end{subfigure}
    \begin{subfigure}[b]{0.5\textwidth}
        \begin{center}
        \includegraphics[width=0.8\textwidth]{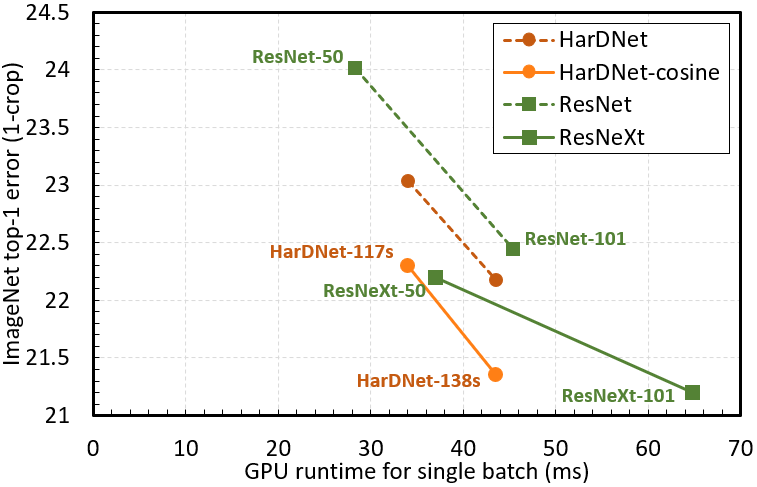}
        \caption{}
        \label{fig:fig5d}
        \end{center}
    \end{subfigure}
    \caption{(a) ImageNet error rate vs. CIO. (b) Error rate vs. model size. (c) Error rate vs. GPU inference time on a single TitanV with torch7. For GPU time of HarDNet-68, please refer to Table \ref{table:table4}. (d) Further comparison among HarDNet-cosine that is trained with cosine learning rate decay and ResNeXt.}
\label{figure:fig5}

\end{figure*}

\subsection{ImageNet Datasets}

To train the six models of HarDNet for the ImageNet classification task, we reuse the torch7 training environment from\cite{K.He, G.Huang} and align all hyperparameters with them. To compare with other advanced CNN architectures such as ResNeXt and MobileNetV2 \cite{mobilenetv2}, we adopt more advanced hyperparameters such as the cosine learning rate decay and a fine-tuned weight decay. The HarDNet-68/39DS models are trained with a batch size of 256, an initial learning rate of 0.05 with cosine learning rate decay, and a weight decay of 6e-5.\\

Investigating the accuracy over CIO, it shows that HarDNet can outperform both ResNet and DenseNet while accuracy over model size is in between them as shown in Fig. \ref{figure:fig5}(a)(b). Fig. \ref{fig:fig5c} shows the GPU inference time results on Nvidia Titan V with torch7, which is quite similar to the trend of Fig. \ref{fig:fig5a} and once again showing the high correlation between CIO and GPU inference time. However, the result also shows that for small models, there is no improvement of GPU inference time for HarDNet compared with ResNet, which we supposed to be due to the number of layers and the concatenation cost. We also argue that, once a discontinuous input tensor can be supported by a convolution operation, the inference time of DenseNet and HarDNet and be further reduced.\\

In Fig. \ref{fig:fig5d}, we compare the state-of-the-art CNN model ResNeXt with our models trained with cosine learning rate decay. Although ResNeXt achieves a significant accuracy improvement with the same model size, there is still an inference time overhead with these models. Since there is no increase of MACs with the ResNeXt, the overhead can be explained by its increase of CIO. \\

In Table \ref{table:table4}, we show the result comparison sorted by CIO for ImageNet, in which HarDNet68/39DS are also included. With the reduced number of layers, the cancel of global dense connections, and the BN-reordering, HarDNet-68 achieves a significant inference time reduction from the ResNet-50. For further comparing CIO between a model using standard convolutions and a model mealy using depth-wise-separable convolutions, we can apply a weighting such as 0.6 on the CIO of the latter. After the weighting, CIO can still be a rough prediction of inference time when comparing among the two very different kinds of model.

\begin{table}[h]
\scalebox{0.8}{
\begin{tabular}{l|c|c|c|c|c|c}
  & \bigcell{c}{Param\\(M)} & \bigcell{c}{MACs\\(B)} & \bigcell{c}{CIO\\(M)}  & \bigcell{c}{GPU\\Time\\(ms)} &\bigcell{c}{Mobile\\GPU\\(ms)} & \bigcell{c}{Top1\\Acc}\\
  \hlineB{5}
  \rowcolor[HTML]{EFFFFF} 
  HarDNet 39DS & 3.5     &	0.44  &	8.2 &	17.8 & 32.5 &		72.1\\
   \hline
  MobileNetV2  &	3.5	 &  0.32  &	13.4 &	23.7& 37.9 &		72.0\\
   \hline
   \rowcolor[HTML]{EFFFFF} 
  HarDNet 68DS &	4.2	 &  0.79  &	15.3 &	31.7& 52.6&		74.3\\
  \hline
  MNetV2 1.4x &	6.1	 &  0.59  &	18.5	&33.0& 57.8 &		74.7\\
  \hlineB{5}
  ResNet 18	   &   11.7  &	1.8	  & 4.7  &	13.0	&-&	69.6\\
  \hline
  SqueezeNet 1.0&	1.2	 &  0.83  &	7.9 &	19.6&-&		60.4\\
  \hline
  \rowcolor[HTML]{EFFFFF} 
  HarDNet 68   &	17.6 &   4.3  &	11.5 &	32.6&-&		76.2\\
  \hline
  \rowcolor[HTML]{EFFFFF} 
  HarDNet 96s  & 9.3	 &   2.5  &	11.7 & 36.4 &-&		74.7\\
  \hline
  \rowcolor[HTML]{EFFFFF} 
  HarDNet 117s &	20.9 &	 4.9  & 16.7 &	57.6&-&		77.0\\
  \hline
  \rowcolor[HTML]{EFFFFF} 
  HarDNet 138s &	35.5 &	 6.7  &	19.6 &	70.5&-&		77.8\\
  \hline
  ResNet 50    &	25   &	 4.1  &	20.7&	46.5	&-&	76.0\\
  \hline
  DenseNet 121 &	7.9	 &   2.9  &	21.9&	51.5&-&		75.0\\
  \hline
  VGG-16       &	138  &	15.5  &	22.6	&79.3	&-&	73.4\\
  \hline
  ResNet 101   & 44.5    &	 7.6  & 30.9	&76.9&-&		77.6\\
  \hline
  DenseNet 201 & 20      &   4.4  &	31.8&	83.9&-&		77.4\\
  \hline
  ResNet 152   & 60.2    &	11.3  &	43.6&	109.7&-&		77.8\\
 \end{tabular}
}
\caption{Test results for ImageNet models, in which GPU time is measured on Nvidia GTX1080 with Pytorch 1.1.0 at 1024x1024 and mobile GPU time is measured on Nvidia Jetson Nano with TensorRT-onnx at 320x320.}
\label{table:table4}
\end{table}

\subsection{Object Detection}
We evaluate HarDNet-68 as a backbone model for a Single Shot Detector (SSD) and train it with PASCAL VOC 2007 and MS COCO datasets. Aligned with the SSD-VGG, we attach an ImageNet-pretrained HarDNet-68 to SSD at the last layers in stride 8 and 16, respectively, and the HDB in stride 32 is discarded. We insert a bridge module after the HDB on stride 16. The bridge module comprises a 3x3 max pooling with stride 1, a 3x3 convolution dilated by 4, and a point-wise convolution, in which both convolutional layers have 640 output channels. We train the model with 300 and 150 epochs for VOC and COCO datasets, respectively. The initial learning rate is 0.004, which is decayed by 0.1 at epochs 60\%, 80\%, 90\% of the total epochs, and the weight decay are 1e-4 and 5e-4 for COCO and VOC, respectively. The results in Table \ref{table:table5} show that our model achieve a similar accuracy with SSD-ResNet101 despite its lower accuracy in ImageNet, which shows the effectiveness of our enhancement on stride 8 with 32 layers that improve the local feature learning for the small-scale objects. Furthermore, HarDNet-68 is much faster than both VGG-16 and ResNet-101, which make it very competitive in real time applications.

\begin{table}[h]
\centering
\scalebox{0.9}{
\begin{tabular}{c|c|c|c}
  & \bigcell{c}{Backbone\\ Model} & \bigcell{c}{VOC 2007\\
mAP} & \bigcell{c}{COCO\\mAP}  \\
  \hlineB{3}
  SSD512& VGG-16 &	79.8 &	28.8 \\
   \hline
  SSD513&	ResNet-101 & 80.6 &	31.2\\
   \hline
  SSD512&	HarDNet-68	& 81.5 & 31.7\\

\end{tabular}
}
\\
\caption{Results in object detection. The comparison data is from \cite{zhang2018single}, with which the training and testing dataset are also aligned.}
\label{table:table5}
\end{table}

%------------------------------------------------------------------------
\section{Discussion}
There is an assumption with the CIO, which is a CNN model that is processed layer by layer without a fusion. In contrast, fused-layer computation for multiple convolutional layers has been proposed \cite{M.Alwani}, in which intermediate layers in a fused-layer group will not produce any memory traffic for feature maps. In this case, the inverted residual module in MobileNetV2 might be a better design to achieve low memory traffic. Furthermore, the depth-wise convolution might be implemented as an element-wise operation right before or after a neighboring layer. In such case, the CIO for depth-wise convolution should be discounted.\\

Results show that CIO still failed to predict the actual inference time in some cases such as comparing two network models with significantly different architectures. As we mentioned before, CIO dominates inference time only when the MoC is below a certain ratio, which is a density of computation within a space of data traffic. In a network model, each of the layers has a different MoC. In some of the layers CIO may dominate, but for the other layers, MACs can still be the key factor if its computational density is relatively higher. To precisely predict the inference latency of a network, we need to breakdown to each of the layers and investigate its MoC to predict the inference latency of the layer.\\

We would like to emphasize the importance of DRAM traffic furthermore. Since the quantization has been widely used for CNN models, both the hardware cost of multiplier and data traffic can be reduced. However, the hardware cost reduction of a multiplier from float32 to int8 is much greater than the reduction of data traffic from the same thing. When developing hardware platform mainly using int8 multipliers, computing power can grow more quickly than the data bandwidth, so data traffic will be even more important in this case. We argue that the best way to achieve the traffic reduction is to increase MoC reasonably for a network model, which might be counter-intuitive to the widely-accepted knowledge of that using more Conv1x1 achieves a higher efficiency. In many cases, we have shown that it is indeed helpful, however.\\

%-------------------------------------------------------------------------

\section{Conclusion}
We have presented a new metric for evaluating a convolutional neural network by estimating its DRAM traffic for feature maps, which is a crucial factor affecting the power consumption of a system. When the density of computation is low, the traffic can dominate inference time more significantly than the model size and operation count. We employ Convolutional Input/Output (CIO) as an approximation of the DRAM traffic, and propose a Harmonic Densely Connected Networks (HarDNet) that achieve a high accuracy-over-CIO and also a high computational efficiency by increasing the density of computation (MACs over CIO).\\

Experiments showed that the proposed connection pattern and channel balancing have made FC-HarDNet to achieve DRAM traffic reduction by 40\% and GPU inference time reduction by 35\% compared with FC-DenseNet. Comparing with DenseNet-264 and ResNet-152, HarDNet-138s achieves the same accuracy with a GPU inference time reduction by 35\%. Comparing with ResNet-50, HarDNet-68 achieves an inference time reduction by 30\%, which is also a desirable backbone model for object detections that enhances the accuracy of a SSD to be higher than using ResNet-101 while the inference time is also significantly reduced from SSD-VGG. In summary, in addition to accuracy-over-model-size and accuracy-over-MACs tradeoffs, we demonstrated that accuracy-over-DRAM-traffic-for-feature-maps is indeed an important consideration when designing neural network architectures.

% \usepackage{multirow}

\section*{Acknowledgement}
This research is supported in part by a grant from the Ministry of Science and Technology (MOST) of Taiwan. We would like to express our gratitude to Taiwan Computing Cloud (TWCC) for providing us with a powerful and stable cloud machine along with top-rated technical support. Without it this research is impossible.

{\small
\bibliographystyle{ieee_fullname}
\bibliography{hardnet}

\begin{thebibliography}{10}\itemsep=-1pt

\bibitem{M.Alwani}
Manoj Alwani, Han Chen, Michael Ferdman, and Peter Milder.
\newblock Fused-layer CNN accelerators.
\newblock In 49th Annual IEEE/ACM International Symposium on Microarchitecture
  (MICRO), pages 1-16, 2016.

\bibitem{segnet}
Vijay Badrinarayanan, Alex Kendall, and Roberto Cipolla.
\newblock SegNet: A Deep Convolutional Encoder-Decoder Architecture for Image
  Segmentation.
\newblock IEEE Transactions on Pattern Analysis and Machine Intelligence,
  39(12): 2481-2495, 2017.

\bibitem{Brostow}
Gabriel~J. Brostow, Julien Fauqueur, and Roberto Cipolla.
\newblock Semantic object classes in video: A high-definition ground truth
  database.
\newblock Pattern Recognition Letters 30 (2), 88-97, 2009.

\bibitem{mem}
Doug Burger, James~R. Goodman, and Alain Kägi.
\newblock Memory bandwidth limitations of future microprocessors.
\newblock In Proceedings of the 23rd annual international symposium on Computer
  architecture (ISCA), Pages 78–89, 1996.

\bibitem{S.Chakradhar}
Srimat Chakradhar, Murugan Sankaradas, Venkata Jakkula, and Srihari Cadambi.
\newblock A dynamically configurable coprocessor for convolutional neural
  networks.
\newblock In International Symposium on Computer Architecture (ISCA), pages
  247-257, 2010.

\bibitem{T.Chen}
Tianshi Chen, Zidong Du, Ninghui Sun, Jia Wang, Chengyong Wu, Yunji Chen, and
  Olivier Temam.
\newblock Diannao: A small-footprint high-throughput accelerator for ubiquitous
  machine-learning.
\newblock In International Conference on Architectural Support for Programming
  Languages and Operating Systems (ASPLOS), pages 269-284, 2014.

\bibitem{Y.Chen}
Yu-Hsin Chen, Joel Emer, and Vivienne Sze.
\newblock Eyeriss: A Spatial Architecture for Energy-Efficient Dataflow for
  Convolutional Neural Networks.
\newblock In International Symposium on Computer Architecture (ISCA), pages
  367-379, 2016.

\bibitem{M.Courbariaux}
Matthieu Courbariaux, Itay Hubara, Daniel Soudry, Ran El-Yaniv, and Yoshua
  Bengio.
\newblock Binarized neural networks: Training Neural Networks with Weights and
  Activations Constrained to +1 or −1.
\newblock arXiv preprint arXiv:1602.02830, 2016.

\bibitem{J.Deng}
Jia Deng, Wei Dong, Richard Socher, Li-Jia Li, Kai Li, and Li Fei-Fei.
\newblock ImageNet: A large-scale hierarchical image database.
\newblock In IEEE Conference on Computer Vision and Pattern Recognition (CVPR),
  pages 248-255, 2009.

\bibitem{M.Drozdzal}
Michal Drozdzal, Eugene Vorontsov, Gabriel Chartrand, Samuel Kadoury, and Chris
  Pal.
\newblock The importance of skip connections in biomedical image segmentation.
\newblock arXiv preprint arXiv:1608.04117, 2016.

\bibitem{A.Aimar}
Alessandro~Aimar et al.
\newblock NullHop: A Flexible Convolutional Neural Network Accelerator Based on
  Sparse Representations of Feature Maps.
\newblock In IEEE Transactions on Neural Networks and Learning Systems, 2018.

\bibitem{pascal}
Mark Everingham, Luc~Van Gool, Christopher K.~I. Williams, John Winn, and
  Andrew Zisserman.
\newblock The PASCAL Visual Object Classes (VOC) Challenge.
\newblock International Journal of Computer Vision, 88(2):303–338, 2010.

\bibitem{C.Farabet}
Cl\'ement Farabet, Berin Martini, Benoit Corda, Polina Akselrod, Eugenio
  Culurciello, and Yann LeCun.
\newblock NeuFlow: A runtime reconfigurable dataflow processor for vision.
\newblock In Proceedings of the IEEE Conference on Computer Vision and Pattern
  Recognition Workshops (CVPRW), pages 109-116, 2011.

\bibitem{P.Gysel}
Philipp Gysel, Mohammad Motamedi, and Soheil Ghiasi.
\newblock Hardware-oriented approximation of convolutional neural networks.
\newblock In International Conference on Learning Representations (ICLR)
  Workshop, 2016.

\bibitem{S.Han}
Song Han, Huizi Mao, and William~J. Dally.
\newblock Deep Compression: Compressing Deep Neural Networks with Pruning,
  Trained Quantization and Huffman Coding.
\newblock In International Conference on Learning Representations (ICLR), 2016.

\bibitem{K.He}
Kaiming He, Xiangyu Zhang, Shaoqing Ren, and Jian Sun.
\newblock Deep residual learning for image recognition.
\newblock In IEEE Conference on Computer Vision and Pattern Recognition (CVPR),
  pages 770-778, 2016.

\bibitem{H.Z.Hu}
Hanzhang Hu, Debadeepta Dey, Allison~Del Giorno, Martial Hebert, and J.~Andrew
  Bagnell.
\newblock Log-DenseNet: How to Sparsify a DenseNet.
\newblock arXiv preprint arXiv:1711.00002, 2017.

\bibitem{G.Huang}
Gao Huang, Zhuang Liu, Laurens van~der Maaten, and Kilian~Q. Weinberger.
\newblock Densely connected convolutional networks.
\newblock In IEEE Conference on Computer Vision and Pattern Recognition (CVPR),
  pages 2261-2269, 2017.

\bibitem{G.Huang2}
Gao Huang, Yu Sun, Zhuang Liu, Daniel Sedra, and Kilian Weinberger.
\newblock Deep networks with stochastic depth.
\newblock In European Conference on Computer Vision (ECCV), pages 646-661,
  2016.

\bibitem{F.N.Iandola}
Forrest~N. Iandola, Song Han, Matthew~W. Moskewicz, Khalid Ashraf, William~J.
  Dally, and Kurt Keutzer.
\newblock Squeezenet: Alexnet-level accuracy with 50x fewer parameters and
  <0.5MB model size.
\newblock arXiv preprint arXiv:1602.07360, 2016.

\bibitem{FCD}
Simon Jégou, Michal Drozdzal, David Vazquez, Adriana Romero, and Yoshua
  Bengio.
\newblock The One Hundred Layers Tiramisu: Fully Convolutional DenseNets for
  Semantic Segmentation.
\newblock In IEEE Conference on Computer Vision and Pattern Recognition
  Workshops (CVPRW), pages 1175-1183, 2017.

\bibitem{A.Krizhevsky}
Alex Krizhevsky.
\newblock Learning multiple layers of features from tiny images.
\newblock Tech Report, 2009.

\bibitem{A.Krizhevsky2}
Alex Krizhevsky, Ilya Sutskever, and Geoffrey~E. Hinton.
\newblock Imagenet classification with deep convolutional neural networks.
\newblock In International Conference on Neural Information Processing Systems
  (NIPS), pages 1097-1105, 2012.

\bibitem{G.Larsson}
Gustav Larsson, Michael Maire, and Gregory Shakhnarovich.
\newblock FractalNet: Ultra-deep neural networks without residuals.
\newblock In International Conference on Learning Representations (ICLR), 2017.

\bibitem{C.Li}
Chao Li, Yi Yang, Min Feng, Srimat Chakradhar, and Huiyang Zhou.
\newblock Optimizing Memory Efficiency for Deep Convolutional Neural Networks
  on GPUs.
\newblock In International Conference for High Performance Computing,
  Networking, Storage and Analysis (SC), pages 633-644, 2016.

\bibitem{coco}
Tsung-Yi Lin, Michael Maire, Serge Belongie, Lubomir Bourdev, Ross Girshick,
  James Hays, Pietro Perona, Deva Ramanan, C.~Lawrence Zitnick, and Piotr
  Doll\'ar.
\newblock Microsoft COCO: Common Objects in Context.
\newblock In European Conference on Computer Vision (ECCV), pages 740–755,
  2014.

\bibitem{J.Long}
Jonathan Long, Evan Shelhamer, and Trevor Darrell.
\newblock Fully convolutional networks for semantic segmentation.
\newblock In IEEE Conference on Computer Vision and Pattern Recognition (CVPR),
  pages 3431-3440, 2015.

\bibitem{D.Miyashita}
Daisuke Miyashita, Edward~H. Lee, and Boris Murmann.
\newblock Convolutional neural networks using logarithmic data representation.
\newblock arXiv preprint arXiv:1603.01025, 2016.

\bibitem{M.Rastegari}
Mohammad Rastegari, Vicente Ordonez, Joseph Redmon, and Ali Farhadi.
\newblock XNOR-net: Imagenet classification using binary convolutional neural
  networks.
\newblock arXiv preprint arXiv:1603.05279, 2016.

\bibitem{samajdar2018scale}
Ananda Samajdar, Yuhao Zhu, Paul Whatmough, Matthew Mattina, and Tushar
  Krishna.
\newblock SCALE-Sim: Systolic CNN Accelerator Simulator.
\newblock arXiv preprint arXiv:1811.02883, 2018.

\bibitem{mobilenetv2}
Mark Sandler, Andrew Howard, Menglong Zhu, Andrey Zhmoginov, and Liang-Chieh
  Chen.
\newblock MobileNetV2: Inverted Residuals and Linear Bottlenecks.
\newblock In IEEE Conference on Computer Vision and Pattern Recognition (CVPR),
  pages 4510–4520, 2018.

\bibitem{K.Simonyan}
Karen Simonyan and Andrew Zisserman.
\newblock Very deep convolutional networks for large-scale image recognition.
\newblock In ICLR, 2014.

\bibitem{R.K.Srivastava}
Rupesh~Kumar Srivastava, Klaus Greff, and Jürgen Schmidhuber.
\newblock Training very deep networks.
\newblock In International Conference on Neural Information Processing Systems
  (NIPS), pages 2377-2385, 2015.

\bibitem{C.Szegedy}
Christian Szegedy, Vincent Vanhoucke, Sergey Ioffe, Jonathon Shlens, and
  Zbigniew Wojna.
\newblock Rethinking the inception architecture for computer vision.
\newblock In IEEE Conference on Computer Vision and Pattern Recognition (CVPR),
  pages 2818–2826, 2016.

\bibitem{zhang2018single}
Shifeng Zhang, Longyin Wen, Xiao Bian, Zhen Lei, and Stan~Z. Li.
\newblock Single-Shot Refinement Neural Network for Object Detection.
\newblock In IEEE Conference on Computer Vision and Pattern Recognition (CVPR),
  pages 4203-4212, 2018.

\bibitem{L.G.Zhu}
Ligeng Zhu, Ruizhi Deng, Michael Maire, Zhiwei Deng, Greg Mori, and Ping Tan.
\newblock Sparsely Aggregated Convolutional Networks.
\newblock In European Conference on Computer Vision (ECCV), 2018.

\end{thebibliography}
}

\end{document}